\documentclass[11pt,a4paper]{article} \usepackage[hyperref]{acl2021}
\usepackage{times}
\usepackage{latexsym}

\usepackage{microtype}
\usepackage[export]{adjustbox}
\usepackage{amsmath}
\usepackage{relsize}
\usepackage{amssymb}
\usepackage{amsfonts}
\usepackage{algorithmic}
\usepackage{graphicx}
\usepackage{textcomp}
\usepackage{xcolor}
\usepackage{todonotes}
\usepackage{booktabs,hhline}
\usepackage{tabu}
\usepackage{multirow}
\usepackage{todonotes}
\usepackage{xargs}                  

\aclfinalcopy
\title{Rule Augmented Unsupervised Constituency Parsing}
\author{Atul Sahay \textsuperscript{1}\thanks{~~Equal contribution}, Anshul Nasery \textsuperscript{1}\footnotemark[1] , Ayush Maheshwari \textsuperscript{1}, \\ \textbf{Ganesh Ramakrishnan \textsuperscript{1} , Rishabh Iyer} \textsuperscript{2} \\
        \textsuperscript{1}Department of CSE, IIT Bombay, India \\ \textsuperscript{2} The University of Texas at Dallas  \\
\texttt{\{atulsahay, anshuln, ayusham, ganesh\}@cse.iitb.ac.in}\\
\texttt{rishabh.iyer@utdallas.edu}
}
\begin{document}
\maketitle
\begin{abstract}
Recently, unsupervised parsing of syntactic trees has gained considerable attention. A prototypical approach to such unsupervised parsing employs reinforcement learning and auto-encoders. However, no mechanism ensures that the learnt model leverages the well-understood language grammar. We propose an approach that utilizes very generic linguistic knowledge of the language present in the form of syntactic rules, thus inducing better syntactic structures. We introduce a novel formulation that takes advantage of the syntactic grammar rules and is independent of the base system. We achieve new state-of-the-art results on two benchmarks datasets, MNLI and WSJ.\footnote{The source code of the paper is available at \href{https://github.com/anshuln/Diora_with_rules}{https://github.com/anshuln/Diora\_with\_rules}. }
\end{abstract}

\section{Introduction}
Syntactic parse trees have demonstrated their importance in several downstream NLP applications such as machine translation~\cite{eriguchi2017learning, zaremoodi2017incorporating}, natural language inference (NLI) \cite{choi2018learning}, relation extraction \cite{gamallo2012dependency} and text classification~\cite{tai2015improved}. Based on linguistic theories that have promoted the usefulness of tree-based representation of natural language text, tree-based models such as Tree-LSTM have been proposed to learn sentence representations~\cite{socher2011semi}. Inspired by the Tree-LSTM based models, many approaches were proposed do not require parse tree supervision~\cite{yogatama2016learning,choi2018learning,Maillard_2019,drozdov2019diora}. However, \citep{williams2018latent, sahay} have shown that these methods cannot learn  meaningful semantics (not even simple grammar), though they perform well on NLI tasks. 
Recently, there has been surge in approaches using weak supervision in the form of rules for various tasks such as sequence classification \cite{safranchik2020weakly}, text classification \cite{chatterjee2020robust, maheshwari}, {\em etc.} These approaches have demonstrated the importance of external knowledge in both unsupervised and supervised setup. To the best of our knowledge, previous works on syntactic parse tree has not utilized such external information. 
In this paper, we propose an approach that leverages linguistic (and potentially domain agnostic) knowledge in the form of explicit syntactic grammar rules while building upon a state of the art, deep and unsupervised inside-outside recursive autoencoder (DIORA; \cite{drozdov2019diora}). DIORA is an unsupervised model that uses inside-outside dynamic programming to compose latent representations from all possible binary trees. We extend DIORA and propose a framework that harness grammar rules to learn constituent parse trees. We use context free grammar (CFG) productions for English language (like  NP \textrightarrow VP NP, PP \textrightarrow IN NP, \textit{etc}) as rules. Note that the construction of such a rule set is a one time effort and our method is independent of any underlying dataset. The rule sets used are available in our github repository.\\
Summarily, our main contributions are : (a) a framework ({\em cf.}, Section~\ref{sec:approach}) that uses (potentially domain agnostic), off-the-shelf CFG to learn to produce constituent parse trees (b) two rule-aware loss functions ({\em cf.}, Section~\ref{sec:loss})  that maximize some form of agreement between the unsupervised model and the rule-based model (c) experimental analysis ({\em cf.}, Section~\ref{sec:expt}), demonstrating improvements on unsupervised constituency parsing  over previous state-of-the art by over \textbf{3}\% on two benchmark datasets.

\section{Background and Related Work}
\label{sec:back}
A brief survey of latent tree learning models is covered in~\cite{williams2018latent}. Several prior works have explored the unsupervised learning of constituency trees \cite{brill1990deducing, ando2000mostly} using dependency parsers~\cite{klein2004corpus} and inside-outside parsing algorithm \cite{drozdov2019diora}. 
Recently, \cite{drozdov2019diora} proposed an unsupervised latent chart tree parsing algorithm, {\em viz.}, DIORA, that uses the inside-outside algorithm for parsing and has an autoencoder-based neural network trained to reconstruct the input sentence.  DIORA is trained end to end using masked language model via word prediction. As of date, DIORA is the state-of-the-art approach to unsupervised constituency parsing.

Exploiting additional semantic and syntactic information that acts as a source of additional guidance rather than the primary objective function has been discussed since 1990s~\cite{sun2017neural}. Recently, \citet{kim2020compound} proposed to learn CFG rules and their probabilities by the parameterizing terminal or non-terminal symbols with neural networks. However, our approach leverages pre-defined language CFG rules and provisions for augmenting an existing (state-of-the-art) inside-outside algorithm with such external knowledge.

 More specifically, we augment DIORA~\cite{drozdov2019diora} with CFG rules to reconstruct the input by exploiting syntactic information of the language.  
We next provide some technical details of the inside-outside algorithm of DIORA.

\begin{figure}
    \centering
    \includegraphics[width=\linewidth]{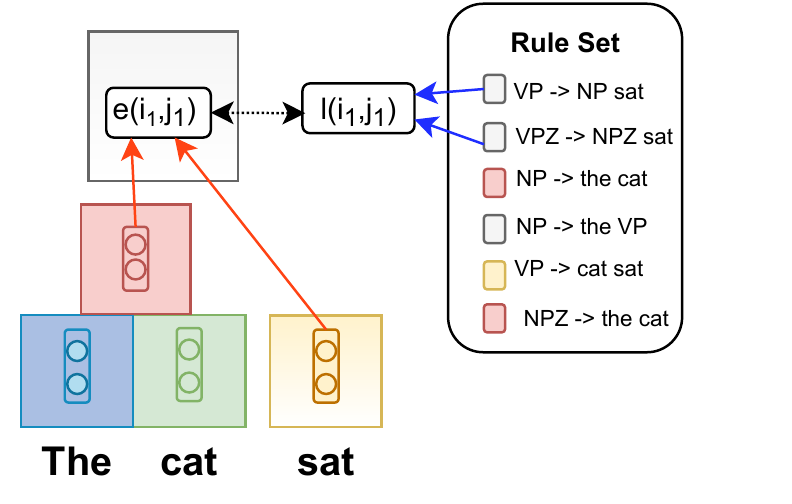}
    \caption{For the input `The cat sat', DIORA computes $e(i_1, j_1)$ compatibility score for each pair of neighboring constituents. $l(i_1,j_1)$ is computed using triggered rules for each span and it interacts with the compatibility score in our loss function as explained in Section~\ref{sec:loss}.}
    \label{fig:arch}
\end{figure}

\subsection{DIORA}
\label{sec:diora}
DIORA learns constituency trees from the raw input text using an unsupervised training procedure that operates like a masked language model or denoising autoencoder. It encodes the entire input sequence into a single vector analogous to the encoding step in an autoencoder. Thereafter, the decoder is trained to reconstruct and reproduce each input word. We next describe the inside and the outside pass of DIORA, respectively. 
\textit{Inside Pass:} Given an input sentence with $T$ tokens $x_0, x_1, x_2 \ldots x_{T-1}$, DIORA computes a \textit{compatibility} score $e$ and a \textit{composition} vector $\overline{a}$ for each pair of neighboring constituents $i$ and $j$. It composes a vector $\overline{a}$ weighing over all possible pairs of constituents of $i$ and $j$:
$
\overline{a}(k) = \sum_{{i,j} \epsilon \{k\} } e(i,j) a(i,j)
$ \& $
\overline{e}(k) = \sum_{{i,j} \epsilon \{k\} } e(i,j) \hat{e}(i,j)
$
The composition vector, $\overline{a}(k)$ is a weighted sum of all possible constituent pairs, ${k}$. Here, $\hat{e}$ is a bilinear function of the vectors from neighboring spans, $\overline{a}(i)$ and $\overline{a}(j)$. Composition vector $\overline{a}(k)$ is learnt using a TreeLSTM or multi-layer neural network (MLP).

\textit{Outside Pass:} The outside pass of DIORA computes an outside vector $\overline{b}(k)$ representing the constituents not in $x_{i:j}$. It computes the values for a target space $(i,j)$ recursively from its sibling $(j+1,k)$ and outside spans $(0,i-1)$ and $(k+1, T-1)$.

\textit{Training and Inference:} DIORA is trained end to end using masked language model via word prediction. The missing token $x_i$ is predicted from the outside vector $\overline{b}(k)$. The training objective uses reconstruction based max-margin loss to predict the original input $x_i$: \\ \\ 
$L_{rec} =  \sum\limits_{i=0}^{T-1} \sum\limits_{i^*=0}^{N-1} \max (0, 1 -\overline{b}(i). \overline{a}(i) + \overline{b}(i). \overline{a}(i^*))
\vspace{0.5cm} $

The chart filling procedure of DIORA is used to extract binary unlabeled parse trees. It uses the CYK algorithm to find the maximal scoring tree in a greedy manner. For each cell of the parse table, the algorithm computes the span $(i,j)$ with the maximal net compatibility score, computed recursively by summing the maximum compatibility score $e(a,b)$ for each constituent of the span.

\section{Our Approach to Rule Augmentation} 
\label{sec:approach}
Our goal is to learn to produce constituency parse trees using input sentences alone and in the absence of ground truth parse trees. We introduce a rule-augmented unsupervised model that leverages generic (potentially domain agnostic) production rules of the language grammar to infer constituency trees. Since most grammar rules for constituency parsing are generic, designing them can be a onetime effort, while being able to leverage their benefits across domains as background knowledge (as we will see in our experiments in Section~\ref{sec:expt}). As described in Section~\ref{sec:diora}, the induction of latent trees in DIORA is based on a CYK-like parsing algorithm that uses the compatibility scores $e(i,j)$ at each cell to merge two constituents in the final tree. We impart supervision through the production rules of English language grammar. 

For each sentence, we associate CFG production rules with constituents $i$ and $j$ in a CYK parse table format. We curate a set of domain-agnostic rules of the form $X \rightarrow Y Z$ and a dictionary of the form $X \rightarrow x$, where $X,Y,Z$ are non-terminals while $x$ is a terminal. Concretely, $X$ represents constituent tags such as S, NP, VP, etc., while $x$ represents words in the vocabulary. 
Using the CYK parsing algorithm on our rule set and each sentence, we first determine which rules are triggered at each cell for a particular sentence. 
Whenever a rule $r$ is triggered for a span $(i,j)$, we weakly associate label $\delta_{(i,j)}(r) = 1$ otherwise 0. We use these weak labels to guide the rule scores $l(i,j)$ for the constituents. Compatibility score observed for the span $(i,j)$ is defined as :
\begin{equation*}
l(i,j) = \frac{\exp\Big(\sum\limits_{p=0}^{P} r_p \delta_{(i,j)}(p) \Big)}{\mathlarger{\sum}\limits_{(a,b) \in \{k\}}\exp\Big(\sum\limits_{p=0}^{P} r_p \delta_{(a,b)}(p) \Big)}
\end{equation*}

where $r_p$ are the \textit{learned} weights associated with each of the production rules and $P$ is the total number of rules. The score sums to $1$ over all spans belonging to a particular cell in the CYK parse table. Intuitively, we aim to align $e(i,j)$ and $l(i,j)$ score to maximize the agreement between model and rules. We note that we use rules only to augment the training objective, and our inference procedure is identical to that of DIORA.

\subsection{Training Objective}
\label{sec:loss}
We learn a  model that minimizes the overall loss $L$ that is  a composition of the reconstruction loss $L_{rec}$ and the rule based agreement loss or $L_{rule}$:
$
L = L_{rec} + \lambda L_{rule}.
$
We propose two alternatives for the loss function $L_{rule}$. \\
\textbf{Cross entropy (CE)} - For each cell $k$ in the CYK parse table, this loss (CE) tries {\em to match the distribution (score)} of $e(i,j)$ induced by DIORA with the distribution $l(i,j)$ induced by the background knowledge: 
\begin{align*}
L_{ce} = \sum_{k} \sum_{(i,j) \in \{k\}} - l(i,j) \log(e(i,j)) 
\end{align*}
\textbf{Ranking Loss (RL)} -
We recall from Section~\ref{sec:diora} that the CYK algorithm finds the maximal scoring tree in a greedy manner based on the highest compatibility score $e(i,j)$ among all spans. Since the final parse tree output by DIORA relies only on the \textit{relative order} of the $e(i,j)$ to decide which span to merge, we propose an alternative rule-based loss that aims {\em to match the relative order} induced by compatibility scores $e(i,j)$ of DIORA at each cell with the order induced by the scores of the rules $l(i,j)$  at that cell. We achieve this through a pairwise ranking loss defined as 
\begin{equation*}
L_{rank} = \sum_{\text{k}} \sum\limits_{\small \substack{(i,j) \\ (i',j')}\in \{\text{k}_{trig}\}}  \bigg( \mathlarger{\Delta}_{\substack{(i,j),\\(i',j')}}^l - \mathlarger{\Delta}_{\substack{(i,j),\\(i',j')}}^e \bigg)^2
\end{equation*}
\normalsize
where $\{\text{k}_{trig}\} = \{(i,j) \in \{k\} | \sum_{p=1}^P \delta_{i,j}(p) \neq 0\}$,  $\Delta_{(i,j),(i',j')}^f = \tanh(f(i,j) - f(i',j'))$ and $p$ is index into the rule set. The set $\{k_{trig}\}$ consists of all spans which have at least one rule triggered in its cell. In cases where our rule set is not extensive enough, we would like our model's compatibility score to rely more on the reconstruction loss, and $\{\text{k}_{trig}\}$ ensures that a sparse rule set does not lead to bad performance. 

\begin{table}[]
\begin{adjustbox}{width=\linewidth}
\begin{tabular}{lccccc}
\toprule
\textbf{Model} &  \multicolumn{1}{c}{\textbf{F1}} & \multicolumn{1}{c}{$\delta$} \\
\midrule
300D Gumbel Tree-LSTM & 25.2 & 4.2\\
~~w/o Leaf GRU  & 29.0& 4.7\\
300D RL-SPINN  & 19.0& 8.6\\
~~w/o Leaf GRU &  18.2&8.6\\
Structural Attentive(Gumble Tree LSTM) &   31.3 & 4.7\\
~~w/o Leaf GRU &   31.0 & 5.3\\
\midrule
300D SPINN & \textbf{74.5$^\dagger$} & 6.2\\
~~w/o Leaf GRU & 65.7$^\dagger$ & 6.4\\

DIORA with PP  & 58.3 & 5.6\\ 
\midrule
Ours: Rule augmented (HR) + RL + PP  & 60.5 & 5.7\\ 
Ours: Rule augmented (HR) + CE + PP  & 59.0 & 5.6 \\
Ours: Rule augmented (AR) + RL + PP & 60.3 & 5.7 \\
Ours: Rule augmented (AR) + CE + PP & \textbf{61.7} & 5.7 \\
\midrule 
\bottomrule
\end{tabular}
\end{adjustbox}
\caption{F1-scores of trees wrt ground truth on the MultiNLI development set. The depth ($\delta$) is the average tree height. All reported numbers are maximum F1-score. PP refers to post-processing heuristic. HR and AR refer to the training rule sets as per Sec \ref{sec:rule_set}. RL and CE refer to the losses from section \ref{sec:loss}. $^\dagger$ indicates scores reported by \cite{williams2018latent} for a fully supervised model}
\label{tab:quantMultiNLI}
\end{table}

\begin{table}[h]

\centering
\begin{adjustbox}{width=\linewidth}
\begin{tabular}{lcccc}
\hline
\multicolumn{1}{c}{\textbf{Model}}  & \multicolumn{1}{c}{\textbf{F1}}  & \multicolumn{1}{c}{\textbf{$\delta$}} \\ \hline
LB& 13.1   & 12.4  \\
RB& 16.5   & 12.4 \\
Random  & 21.4   & 5.3 \\
Balanced  & 21.3 & 4.6 \\
RL-SPINN \cite{choi2018learning} & 13.2 & - \\
ST-Gumbel - GRU   \cite{yogatama2016learning}& 22.8 ±1.6 & -   \\ \hline
PRPN-UP \cite{shen2018neural} & 38.3 & 5.9  \\
PRPN-LM& 35.0 & 6.2 \\
ON-LSTM \cite{shen2018ordered}& 47.7 & 5.6 \\
DIORA& 48.9 & 8.0 \\ \hline
PRPN-UP+PP & 45.2 & 6.7 \\
PRPN-LM+PP & 42.4 & 6.3 \\
DIORA+PP & 55.7 & 8.5 \\ \hline
Neural PCFG \cite{kim2020pretrained} $^*$ & 50.8 & - \\
Compound PCFG \cite{kim2020compound}$^*$
& 55.2 & -\\ 
300D SPINN & \textbf{59.6$^\dagger$} & -\\ \hline

Ours: Rule augmented (HR)
+ RL + PP  & 56.5 & 7.1\\ 
Ours: Rule augmented (HR)
+ CE + PP  & 55.3 & 7.2 \\
Ours: Rule augmented (AR) 
+ RL + PP & 55.9 & 7.1 \\
Ours: Rule augmented (AR)
+ CE + PP & \textbf{58.3} & 7.3 \\ \hline \hline
\end{tabular}
\end{adjustbox}
\caption{ Performance on WSJ test set for binary constituency parsing including punctuation characters. HR and AR refers to handcrafted and automated rules respectively. RL and CE are rule loss and cross entropy loss respectively. ($\delta$) is the average tree height. PP refers to post-processing heuristic. $^\dagger$ indicates scores reported by \cite{williams2018latent} for a fully supervised model. $^*$ are reported by \cite{kim2020pretrained}.}
\label{tab:wsj}
\end{table}
   
\section{Experiments}
\label{sec:expt}
We evaluate our rule augmented model and compare it against baselines on the tasks of unsupervised parsing, unsupervised segment recall, and phrase similarity.

\subsection{Data}
We evaluate our model on two data sets: The Wall Street Journal (WSJ) and MultiNLI. WSJ is an extraction of PennTree Bank~\cite{article} containing human-annotated constituency parse trees. MultiNLI consists of Stanford generated parse trees ~\cite{manning2014stanford} as the ground truth. MultiNLI is originally designed for evaluating NLI tasks, but is often also utilized to evaluate constituency parse trees. We train on the complete NLI dataset, which is a composition of the MultiNLI and SNLI train sets. We evaluate model performance on the MultiNLI dev set and WSJ test set (split 23) following the experimental setting and evaluation metrics in~\cite{drozdov2019diora}. Further details are provided in the appendix.
We initialize our model with the trained weights of DIORA and evaluate on unsupervised constituency parsing and segment recall. We also perform the post-processing (PP) of generated trees by attaching the trailing punctuation to the root node, exactly as carried out by~\cite{drozdov2019diora}. 

\subsection{Rule Set}
\label{sec:rule_set}
We consider two rule-sets: (i) {\bf Set of Handcrafted Rules (HR)} consists of   2500 human created CNF production rules ii) To assess robustness of the rule-augmented method to the preciseness of the rule set, we present comparison by instead using a {\bf set of Automated Rules' (AR)} which consists of the 2500 most frequently occurring CNF production rules extracted from the trees of automatically (using the Stanford CoreNLP parser) parsed SNLI corpus. Further details about these rule sets can be found in the appendix.
We also use a train-set specific dictionary containing the POS (part-of-speech) tags of words in the training vocabulary for the terminal CFG productions for CYK parsing.

\begin{table*}[!h]
\begin{tabular}{l|ccclll}
\hline
\multicolumn{1}{c|}{\textbf{Model}} & \textbf{SBAR}              & \textbf{NP}                         & \textbf{VP}                         & \textbf{PP}     & \textbf{ADJP}   & \textbf{ADVP}   \\ \hline
LB $\dagger$                                 & 5\%                        & 11\%                                & 0\%                                 & 5\%             & 2\%             & 8\%             \\
RB   $\dagger$                               & 68\%                       & 24\%                                & 71\%                                & 42\%            & 27\%            & 38\%            \\
Random $\dagger$                             & 8\%                        & 23\%                                & 12\%                                & 18\%            & 23\%            & 28\%            \\
Balanced $\dagger$                           & 7\%                        & 27\%                                & 8\%                                 & 18\%            & 27\%            & 25\%            \\ \hline
PRPN-UP \cite{shen2018neural}                              & 55.4\%                     & 59.8\%                              & 31.6\%                              & 60.2\% & 36.0\%          & 50\%            \\
PRPN-LM                             & 40.3\%                     & 68.7\%                              & 39.3\%                              & 49.7\%          & 34.2\%          & 39.2\%          \\
DIORA                                & 61.3\%                     & 76.7\%                              & 62.8\%                              & 59.5\%          & 60.4\%          & 69.3\%          \\ \hline
PRPN  (tuned)$\dagger$                         & 50\%                       & 59\%                                & 46\%                                & 57\%            & 44\%            & 32\%            \\
ON  (tuned) \cite{shen2018ordered}                           & 51\%                       & 64\%                                & 41\%                                & 54\%            & 38\%            & 31\%            \\
Neural PCFG \cite{kim2020pretrained}                         & 52\%                       & 71\%                                & 33\%                                & 58\%            & 32\%            & 45\%            \\
Compound PCFG \cite{kim2020compound}                        & \multicolumn{1}{l}{56\%}   & \multicolumn{1}{l}{74\%}            & \multicolumn{1}{l}{41\%}            & 68\%            & 40\%            & 52\%            \\ \hline
Ours: Rule augmented (HR)+ RL       & \textbf{71.1\%}            & 77.2\%                              & 65.8\%                              & 59.4\%          & \textbf{62.9\%} & 69.5\%          \\
Ours: Rule augmented (HR)+ CE       & \multicolumn{1}{l}{68.3\%} & \multicolumn{1}{l}{75.4\%}          & \multicolumn{1}{l}{66.5\%}          & \textbf{60.5\%} & 61\%            & \textbf{70.8\%} \\
Ours: Rule augmented (AR)+ RL      & \multicolumn{1}{l}{71\%}   & \multicolumn{1}{l}{76.4\%} & \multicolumn{1}{l}{\textbf{69.1\%}} & 58.6\%          & 61\%            & 64.8\%          \\
Ours: Rule augmented (AR)+ CE      & \multicolumn{1}{l}{70\%}   & \multicolumn{1}{l}{\textbf{77.5\%}}          & \multicolumn{1}{l}{67\%}            & 58.6\%          & \textbf{62.2\%} & 70\%            \\ \hline \hline
\end{tabular}
\caption{Segment recall from WSJ by phrase type; $\dagger$ are reported by~\citet{kim2020pretrained}. } 
\label{tab:segment}

\end{table*}


\begin{table*}
\centering
\begin{adjustbox}{width=0.7\linewidth}

\begin{tabular}{l|ccc|cc}
\hline
\multicolumn{1}{c|}{\textbf{Model}}                      & \multicolumn{3}{c|}{\textbf{CONLL 2000}}                                                                   & \multicolumn{2}{c}{\textbf{CONLL 2012}}                              \\
& \multicolumn{1}{l}{\textbf{P@1}} & \multicolumn{1}{l}{\textbf{P@10}} & \multicolumn{1}{l|}{\textbf{P@100}} & \multicolumn{1}{l}{\textbf{P@1}} & \multicolumn{1}{l}{\textbf{P@10}} \\ \hline
DIORA                          & 0.974                     & 0.969                      & 0.943                        & 0.815                     & 0.759                      \\
Ours: Rule augmented(AR)+ RL & 0.976                      & 0.968                      & 0.941                        & 0.813                     & 0.755                      \\
Ours: Rule augmented(HR)+ CE  & \textbf{0.978}             & \textbf{0.97}             & 0.941                        & 0.786                     & 0.717                      \\
Ours: Rule augmented(AR)+ CE & 0.970                     & 0.965                      & 0.937                        & 0.809                     & 0.745                      \\
Ours: Rule augmented(HR)+ RL  & 0.976                      & 0.970                      & \textbf{0.944}               & \textbf{0.824}            & \textbf{0.760}             \\ \hline
\end{tabular}
\end{adjustbox}
\caption{Phrase similarity scores on CoNLL2000 and CoNLL 2012 tasks.} 
\label{tab:similarity}
\end{table*}

\subsection{Unsupervised  Parsing}

In Tables~\ref{tab:quantMultiNLI} and \ref{tab:wsj}, we present comparison between different approaches on the MultiNLI dev set and WSJ test set. We observe that our rule augmented approach outperforms the state of the art with respect to the max-F1 score. registering a maximum increase of 3.4  and 3.1 F1 points over DIORA respectively. 
The HR trained models outperform DIORA on both datasets, demonstrating that rule creation is indeed a one-time process and independent of domain. 
We also report parsing scores of a fully supervised model SPINN from \cite{williams2018latent} as an upper bound, and RL-SPINN \cite{choi2018learning}, a distantly supervised model.
%

\subsection{Constituency Segment Recall}
In Table \ref{tab:segment}, we present the breakdown of constituent recall across the 6 most common types. Our approach achieves the highest recall across all the types and is the only model to perform effectively on SBAR and NP. Unlike other approaches, our approach consistently close to or the best recall score. \\
We observe that rule augmentation using HR is more beneficial than AR with respect to precise evaluation measures such as Constituency, Segment Recall and Phrase Recall but yields smaller improvements than AR with respect to looser evaluation measures such as max F1 of  Unsupervised Parsing. This can be possibly attributed to our observation that the extracted (most frequent) rules from SNLI, have (around 25\%) higher coverage on the training set than HR, but appear to be semantically less precise. 

\subsection{Phrase Similarity}
We also employed the phrase similarity strategy followed by ~\cite{drozdov2019diora}. Phrase Similarity scores measures the models capability to learn meaningful representation for spans of the text. Generally, most models focus more on generating the tokens representation and then use some ad-hoc arithmetic operations to generate representation for the larger spans of text thus losing the essence of the context that ties the words of the span. 

To evaluate on the phrase similarity task we consider two data sets of labeled phrases: 1) CoNLL 2000 ~\cite{tjong2000introduction}, which is a shallow parsed dataset and contains spans of verb phrases, noun phrases, preposition phrases \emph{etc.}, and 2) CoNLL 2012 ~\cite{pradhan2012conll} which  is a named entity dataset containing 19 different entity types. For the evaluation routine, we first generated the phrase representation of labeled spans whose length is greater than one. Cosine similarity is then used to obtain the similarity score of it with respect to all other labeled spans. We then calculate if the label for that query span matches the labels for each of
the K most similar other spans in the dataset.In Table \ref{tab:similarity} we report precision@K for both datasets and various values of K. The baseline numbers are reported using the weights of DIORA provided by the authors.
\section{Conclusion}
In this work, we leverage linguistically grounded and domain agnostic CFG rules for language to induce parse trees and representations of constituent spans. We show that our approach augmented with generic, linguistically grounded grammatical rules, is easily able to outperform previous methods on constituency parsing and obtain higher segment recall. 

\section*{Acknowledgements}
We thank anonymous reviewers for providing constructive feedback. Ayush Maheshwari is supported by a Fellowship from Ekal Foundation (www.ekal.org). We are also grateful to IBM Research, India (specifically the IBM AI Horizon Networks - IIT Bombay initiative) for their support and sponsorship.
\bibliographystyle{acl_natbib}
\bibliography{refs.bib} 

\clearpage
\appendix


\section*{Appendix}
\section{Training and evaluation details}
We use the multi-layered neural network model provided by the authors of \cite{drozdov2019diora} to initialize our model's weights. We have a learning rate of $10^{-4}$ and a batch size of $32$. Using these settings we train our models for $500000$ steps. We use weights $\lambda_{ranking} = 10^{-1}$ and $\lambda_{ce} = 1.0$ for the rule based losses. All other model parameters are same as the ones set in \cite{drozdov2019diora}. We run all our experiments on Nvidia RTX 2080Ti GPUs 12 GB RAM over Intel Xeon Gold 5120 CPU having 56 cores and 256 GB RAM. It takes about 2 days to train the model on NLI data. \\
For evaluation, we have reported the tree F-1 score for MNLI dev and WSJ test set. The metric computes the F-1 score for each tree based on the constituent spans induced in the predicted tree against the constituent spans in the ground truth. We further binarize the WSJ test set using the Stanford CoreNLP Parser and report scores on unlabelled binary trees. \\
We find that training with AR helps us achieve better results on both MNLI as well as on WSJ. This could be because extracted rules from SNLI have wider coverage on the training set than HR resulting in a stronger training signal and better performance. Further, our ranking loss performs better for HR extracted rules, indicating its efficacy with non-extensive rule sets, i.e. in the cases where the training signal is not rich. In such cases when some cells may not have any triggering rules, the ranking loss ensures that the model's decision is guided by the reconstruction loss.  \\
We also find that the generic background knowledge of English grammar (HR) helps the model to better chunk constituents that are rarer (e.g. SBAR), while dataset-specific rules (AR) might benefit its overall tree structures more, leading to higher unlabelled F1 scores.


\section{Rule sets}
In this work we utilise two distinct rule sets - (i) The first rule set (HR) consists of   2500 human created CNF production rules ii) the other set (AR) consisted of 2500 most frequently occurring CNF production rules extracted from the trees of automatically parsed SNLI corpus. All rules in both these sets consist only of non-terminals. The rules in (HR) come from observing human annotated parse trees from the PTB train set and consists of 2500 rules in the Chomsky Normal Form. The rules in (AR) are programatically extracted from the parse trees generated by running the Stanford Parser on the SNLI train set. We only retain the 2500 most frequently occurring productions from the set to match the size of the HR set. We note however that these rules have a higher coverage on the train data. We also provides \texttt{rules-AR.txt} and \texttt{rules-HR.txt} in the github repository. \\ 
In our training procedure, we aim to learn a weight $r_p$ for each production rule $p$ in our train set. Table \ref{tab:top_rules} shows the top 10 most important (i.e. the ones with the highest $r_p$) rules from the grammar as determined by our models.
\begin{table*}[]
\begin{tabular}{@{}lllll@{}}
\toprule
\textbf{Model}             & \textbf{Top Rules}                                                                                                                                     &  &  &  \\ \midrule
AR ranking loss & \begin{tabular}[c]{@{}l@{}}NP ---\textgreater PRP NP$|$CD-JJ-VBN-NNS\textgreater \\ NP ---\textgreater PRP NP$|$\textless{}NNP-NNP-NN\textgreater \\ S ---\textgreater S S$|$\textless{}CC-SINV-.\textgreater \\ ADVP ---\textgreater IN ADVP$|$\textless{}CC-JJ\textgreater \\ VP ---\textgreater VBN VP$|$\textless{}``-NP-''-PP\textgreater \\ NP ---\textgreater NP NP$|$\textless{}NN-NN-''\textgreater \\ PP ---\textgreater IN , \\ NP ---\textgreater NP NP$|$\textless{}ADJP-PP-SBAR\textgreater \\ NP ---\textgreater NP NP$|$\textless{}NNS-S\textgreater \\ PP ---\textgreater `` PP$|$\textless{}IN-NP\textgreater{}\end{tabular}                                                                                       &  &  &  \\
\hline
HR cross entropy     & \begin{tabular}[c]{@{}l@{}}NP ---\textgreater CD NNS \\ S-CLR ---\textgreater VP \\ PP-PRP ---\textgreater IN NP \\ QP$|$\textless{}CD-TO-CD-CD\textgreater ---\textgreater CD QP$|$\textless{}TO-CD-CD\textgreater \\ S-2 ---\textgreater NP S-2$|$\textless{}VP-.\textgreater \\ VP ---\textgreater VB VP$|$\textless{}NP-ADVP-S\textgreater \\ NP$|$\textless{},-''-SBAR\textgreater ---\textgreater , NP$|$\textless{}''-SBAR\textgreater \\ S ---\textgreater NP S$|$\textless{}NP-VP-.\textgreater \\ NP$|$\textless{}JJS-NNS\textgreater ---\textgreater JJS NNS \\ S-2$|$\textless{}VP-.\textgreater ---\textgreater VP .\end{tabular}                                                                                                  &  &  &  \\
\hline
HR ranking loss       & \begin{tabular}[c]{@{}l@{}}NP ---\textgreater PRP NP$|$\textless{}NNP-CD-NN\textgreater \\ NP ---\textgreater DT NP$|$\textless{}JJ-NN-NN\textgreater \\ S$|$\textless{}NP-VP-.---\textgreater ---\textgreater NP S$|$\textless{}VP-.---\textgreater \\ VP$|$\textless{}NP-S\textgreater ---\textgreater NP S \\ NP$|$\textless{}NNP-NNP-NNP-NN-NN\textgreater ---\textgreater NNP NP$|$\textless{}NNP-NNP-NN-NN\textgreater \\ NP ---\textgreater JJ NP$|$\textless{}NN-POS\textgreater \\ VP$|$\textless{}CC-,-VP\textgreater ---\textgreater CC VP$|$\textless{},-VP\textgreater \\ NP ---\textgreater NNP NP$|$\textless{}CC-NNS\textgreater \\ NP$|$\textless{}:-SBAR\textgreater ---\textgreater : SBAR \\ ADJP ---\textgreater \$ CD\end{tabular}  &  &  &  \\
\hline
AR cross entropy    & \begin{tabular}[c]{@{}l@{}}NP ---\textgreater DT NP$|$\textless{}JJ-NNP-NNP-POS\textgreater \\ VP ---\textgreater VB VP$|$\textless{}NP-PRT\textgreater \\ S$|$\textless{}S-:\textgreater ---\textgreater S : \\ VP ---\textgreater VBZ VP \\ PRN ---\textgreater , PRN$|$\textless{}CC-PP\textgreater \\ VP$|$\textless{}CC-VBG-NP-PP\textgreater ---\textgreater CC VP$|$\textless{}VBG-NP-PP\textgreater \\ NP$|$\textless{},-NP-,-VP-.\textgreater ---\textgreater , NP$|$\textless{}NP-,-VP-.\textgreater \\ S$|$\textless{}PP-,-VP-.\textgreater ---\textgreater PP S$|$\textless{},-VP-.\textgreater \\ PP ---\textgreater RB PP$|$\textless{}CC-RB-NP\textgreater \\ NP ---\textgreater JJ NP$|$\textless{}NNP-NNP\textgreater{}\end{tabular} &  &  &  \\ \bottomrule
\end{tabular}
\caption{Rules with the highest weights as learnt by our models } 
\label{tab:top_rules}

\end{table*}

\begin{figure*}[!t]
\begin{tabular}{| c |@{}c@{} | @{}c@{}|}
\hline

Our Results &
{\includegraphics[height=50mm,width=50mm,valign=b]{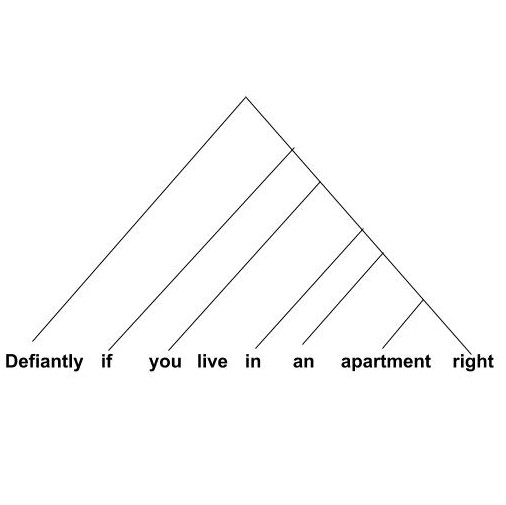}} 
&
{\includegraphics[height=60mm, width=65mm,valign=b]{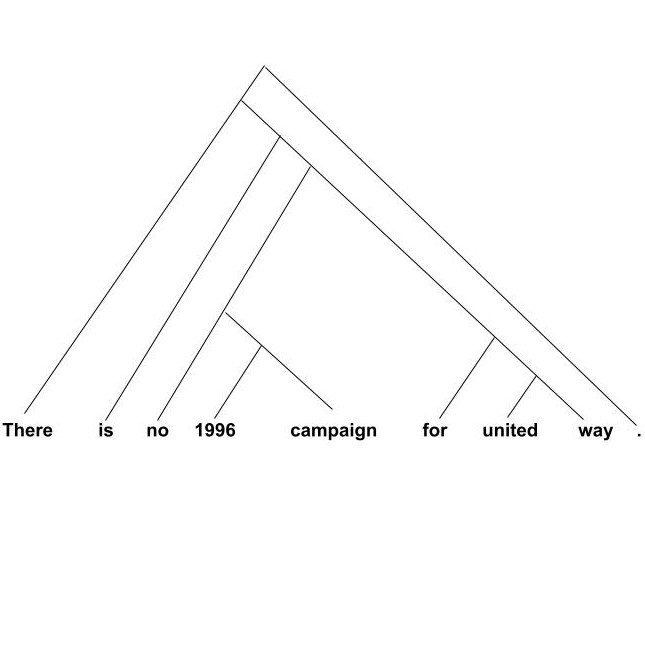}}
\\
\hline
DIORA's Results &
{\includegraphics[width=50mm]{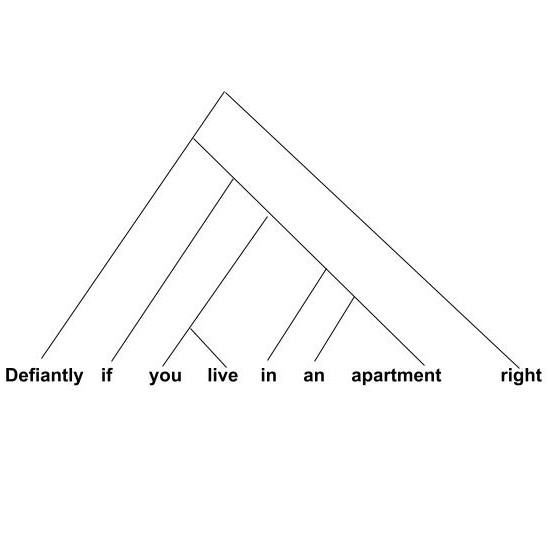}}
&
{\includegraphics[width=60mm]{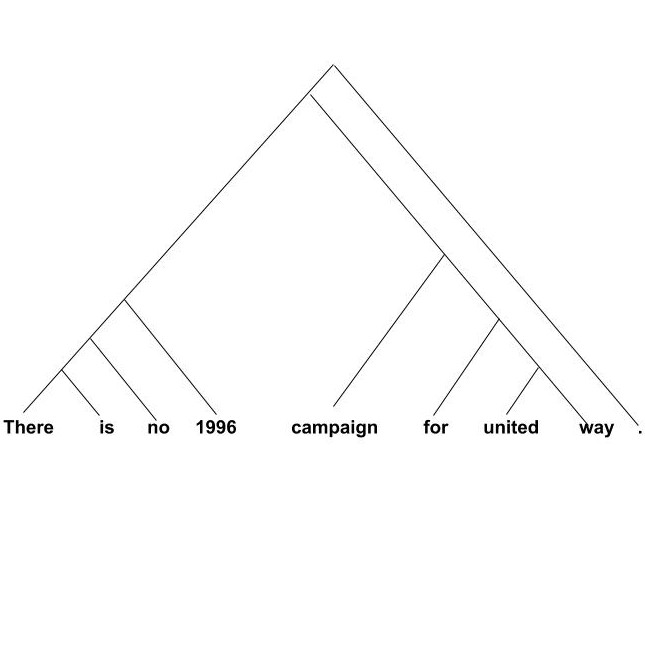}}
\\
\hline
Ground Truth &
{\includegraphics[width=50mm]{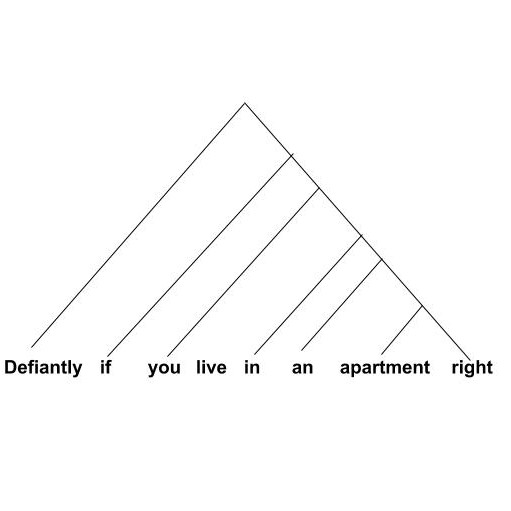}}
 &
{\includegraphics[width=60mm]{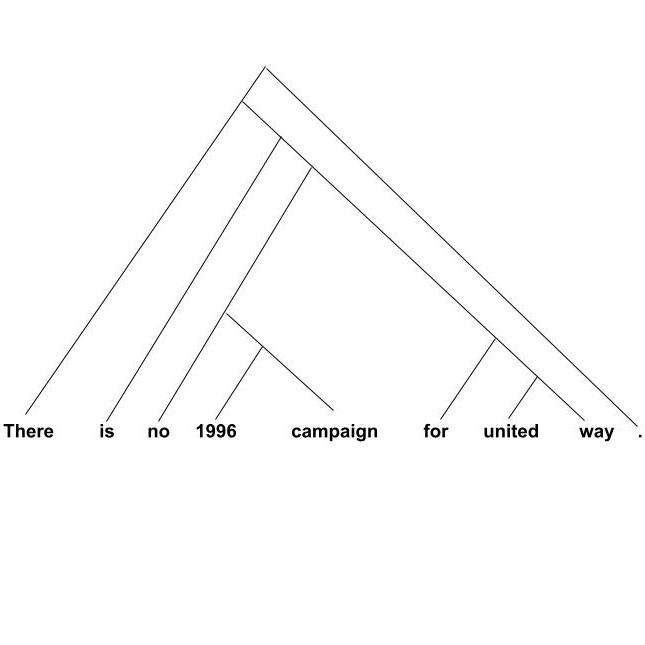}}
  \\
\hline
\end{tabular}

\caption{Comparison of induced trees by our model and DIORA with the ground truth trees}
\label{fig:trees}
\end{figure*}


\section{Learnt Trees}
In this section we present examples of trees induced by DIORA and our model. The first row of fig. 1 shows an example where our tree matches the ground truth exactly while DIORA does not, and the second row of fig. 1 shows an example where both models do not provide exact matches, but our model is able to capture the syntax better.

\end{document}